\documentclass{article}
\usepackage{iclr2027_conference,times}

\usepackage[utf8]{inputenc}
\usepackage[T1]{fontenc}
\usepackage{hyperref}
\usepackage{url}
\usepackage{booktabs}
\usepackage{amsmath}
\usepackage{amssymb}
\usepackage{graphicx}
\usepackage{xcolor}
\usepackage{colortbl}
\usepackage{colortbl}
\usepackage{xspace}
\usepackage{pifont}
\usepackage{microtype}
\usepackage{multirow}
\usepackage{float}
\usepackage{needspace}
\usepackage{placeins}
\usepackage{caption}
\usepackage{placeins}

\newcommand{\method}{\textbf{ACS}\xspace}

\title{The Model Knows When to Stop: Training-Free Early Stopping for Long-Context Reading}

 \iclrfinalcopy  
\author{%
\makebox[\textwidth][c]{%
\begin{tabular}{@{}c@{\hspace{2.2em}}c@{\hspace{2.2em}}c@{}}
Muath Alyobi$^{1}$ &
Mohamed Eltahir$^{1}$ &
Almoayyad Abuljdail$^{1}$ \\[0.25em]
Riyadh Almutawa$^{1}$ &
Tanveer Hussain$^{2\ddagger}$ &
Naeemullah Khan$^{1\S}$
\end{tabular}}\\[1.0em]
\makebox[\textwidth][c]{$^{1}$King Abdullah University of Science and Technology (KAUST), Thuwal, Saudi Arabia}\\
\makebox[\textwidth][c]{$^{2}$Department of Computer Science, Edge Hill University, Ormskirk, England}\\[0.3em]
\makebox[\textwidth][c]{\small\texttt{\{muath.alyobi, mohamed.hamid, almoayyad.abuljdail,}}\\
\makebox[\textwidth][c]{\small\texttt{ \{riyadh.almutawa, naeemullah.khan\}@kaust.edu.sa}}\\
\makebox[\textwidth][c]{\small\texttt{hussaint@edgehill.ac.uk}}
}

\begin{document}

\vspace*{-30pt}
\maketitle
 \lhead{Preprint.}

 {\renewcommand{\thefootnote}{\fnsymbol{footnote}}
 \footnotetext[3]{Corresponding author.}
 \footnotetext[4]{Principal Investigator (PI).}
 \footnotetext{Code: https://github.com/mu18th/Answer-Convergence-Stopping}
 }

 \vspace{-18pt}

\begin{figure}[H]
    \centering
    \makebox[\linewidth][c]{%
        \includegraphics[width=1.14\linewidth]{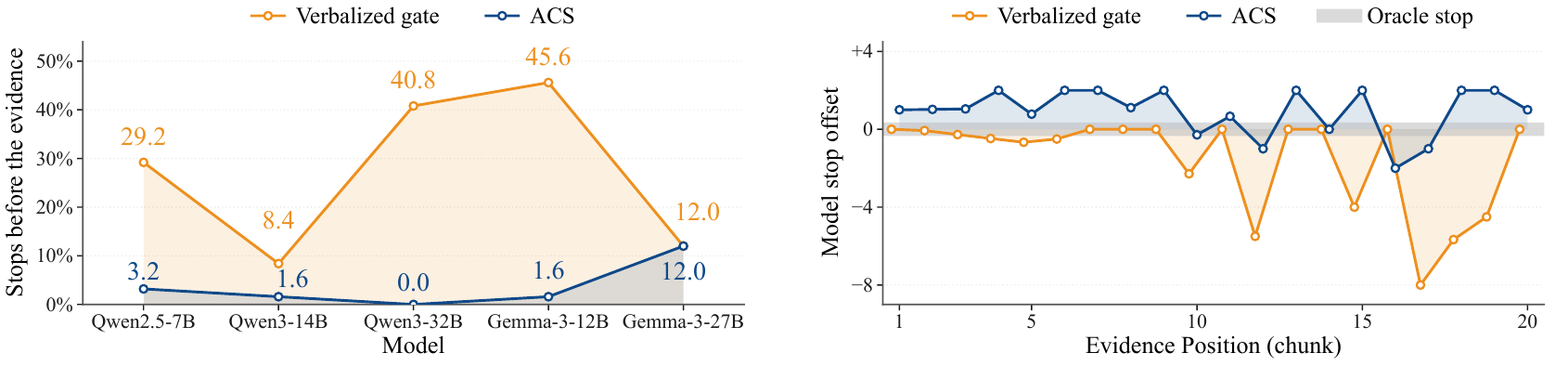}
    }
    \vspace{-2mm}
    \caption{Evidence-aligned stopping on S-NIAH. \textbf{Left:} cross-model premature-stop behavior under the same shared setting across five models. \textbf{Right:} within-run stopping behavior relative to the evidence position. The oracle stops exactly at the evidence-containing chunk, so zero denotes the oracle stop, positive values indicate over-reading, and negative values indicate premature stopping.}
    \label{fig:consistency}
\end{figure}

 \vspace{-4pt}
\begin{abstract}
Language models often process long inputs sequentially in chunks, but continuing to read after sufficient evidence has been acquired wastes computation. Existing stopping mechanisms either learn sufficiency from internal activations or train an exit gate, while a simpler alternative asks the model whether it has read enough. We introduce \textbf{Answer-Convergence Stopping} (\method{}), a training-free stopping rule that measures rather than asks. After each chunk, it probes the frozen model's current answer state and stops when that state is both confident and stable. The rule requires only output-side generation and token log probabilities, has no trained components, and uses one shared configuration across models and benchmarks. Because a stopping policy can save computation simply by stopping too early, we evaluate the stopping decision itself using evidence position where available. On the full LongBench-v2 with two frontier models, \method{} is the only stopping policy that matches or exceeds full-reading accuracy. Furthermore, across 250 S-NIAH questions, the premature stopping rate for \method{} across five models from two families ranges from 0\% to 12\%, compared to 8.4\% to 45.6\% for the verbalized gate. Taken together, \method{} reveals that by properly utilizing the output signals of frozen models, we can achieve favorable behaviors like adaptive stopping without the need for additional training. 
\end{abstract}

\section{Introduction}\label{sec:intro}

Language models are increasingly used for tasks involving long inputs that exceed their context window. A common strategy is to split the input into chunks and process them sequentially, carrying forward information from earlier chunks rather than presenting the full context in a single call. This pattern underlies agent-based readers~\citep{zhang2024chainofagents}, memory agents trained with reinforcement learning~\citep{yu2026memagent, sheng2026grumem}, and programmatic runtimes that recursively decompose long inputs~\citep{zhang2025rlm, roy2026lambdarlm}. However, incremental processing still incurs substantial cumulative cost and does not by itself determine when enough has been read. As GRU-Mem~\citep{sheng2026grumem} puts it, the loop ``lacks an exit mechanism, leading to unnecessary computation after even sufficient evidence is collected''.

Recent work has therefore explored how to determine when enough context has been processed. Dynamic context cutoff~\citep{xie2025knowing} finds that specific attention heads encode sufficiency and monitors them with lightweight classifiers trained on labels derived from ground-truth answer locations. This provides a strong learned sufficiency signal, but requires access to internal activations and training a classifier from sufficiency-labeled examples. GRU-Mem~\citep{sheng2026grumem} adds an exit gate to a memory agent and trains it through an explicit exit reward. DCC~\citep{xie2025knowing} also considers a training-free alternative: directly prompting the model to judge whether the accumulated context is sufficient. Their results show a clear dependence on model scale: self-prompting improves as model size increases, while activation-based probes remain stronger across the evaluated models.

\begin{figure}[t]
\centering
\includegraphics[width=\linewidth]{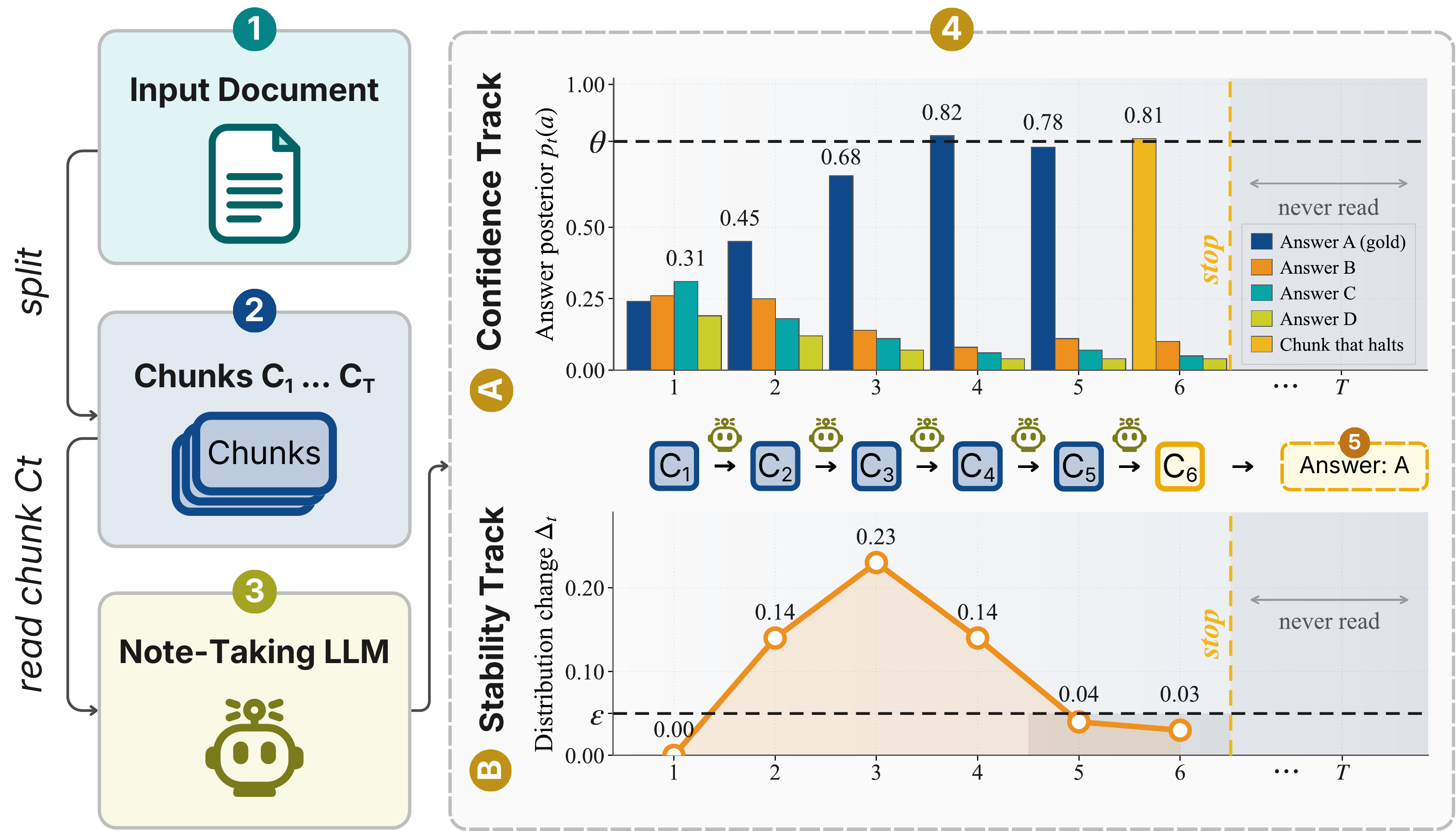}
\caption{
Overview of \method{}. The document is processed sequentially in chunks while
a frozen model maintains running notes. After each chunk, a separate probe
produces the current answer state $b_t$ and confidence $c_t$, and consecutive
answer states define the change $\delta_t$. The stopping rule halts at the first
step where $c_t \geq \theta$ and the mean change $\Delta_t$ over the last
$\min(t, w)$ answer states satisfies $\Delta_t \leq \varepsilon$.
}
\label{fig:overview}
\end{figure}

Rather than asking the model to judge whether it has read enough or probing its internal activations, we measure how its answer state evolves as more context is read. After each chunk, we probe the frozen model’s current answer state. We hypothesize that this evolving answer state is sufficiently informative to serve as a signal for when further reading is no longer necessary. We call this a \emph{measured} signal, in contrast to \emph{latent} signals extracted from internal activations by trained probes and \emph{asked} signals obtained through self-report. \method{} (Figure~\ref{fig:overview}) turns this signal into a stopping rule with two criteria. The confidence criterion measures whether the belief is sufficiently committed. The stability criterion measures whether that belief has stopped changing. The second criterion is important because confidence alone can be transient. Our default evaluation uses one shared conservative configuration across models and benchmarks, while we separately study the confidence threshold as an accuracy-efficiency operating point. No stopping component is trained.

A stopping rule should be judged on where it stops, not only on the cost it saves, because a policy can save by stopping before the evidence and guessing. On needle benchmarks, the evidence position is known by construction, providing an explicit reference for whether a stopping decision occurs before or after the required evidence. We use this information only for evaluation, reporting premature-stop rate, over-read, accuracy regret, and savings capture relative to an oracle that stops at the evidence. This allows us to distinguish evidence-aligned savings from savings obtained simply by truncating the reading process early.

Figure~\ref{fig:consistency} summarizes both the within-run and cross-model stopping behavior. On 250 S-NIAH instances with Qwen3-14B~\citep{qwen3}, the right panel shows that \method{} stops before the evidence on only \(1.6\%\) of questions, compared with \(8.4\%\) for the verbalized gate, while still capturing \(60.7\%\) of the savings available to an oracle that stops exactly at the evidence. The left panel shows that this behavior extends across models: premature stopping for \method{} remains between \(0.0\%\) and \(12.0\%\) across five models, while the verbalized gate ranges from \(8.4\%\) to \(45.6\%\). Together, the two views show that \method{} produces evidence-aligned stopping while maintaining low premature-stop rates across models.

\textbf{Contributions.}
(1) A training-free stopping rule for chunked reading, defined for both option-constrained and open-ended answers, using only output-side generation and log probabilities.
(2) An evidence-aware evaluation protocol that measures stopping decisions against known evidence position using premature-stop rate, over-read, oracle-normalized regret, and savings capture.
(3) A cross-setting evaluation showing that the same confidence-stability formulation remains effective across model families, scales, benchmarks, and answer formats, with a shared conservative default and separately studied accuracy-efficiency operating points.

\section{Related Work}\label{sec:related}


\textbf{Reading in pieces.} Long-context systems often process inputs incrementally, either by carrying information forward across segments or by selectively decomposing and navigating the input. Chain-of-Agents~\citep{zhang2024chainofagents} assigns successive text segments to worker agents that pass a communication unit forward, followed by a manager that produces the final answer. MemAgent~\citep{yu2026memagent} maintains a fixed-length memory that is overwritten after each segment and trains the memory-update policy with reinforcement learning, yielding linear processing complexity. RLM~\citep{zhang2025rlm} lets the model write code to inspect and decompose the input, whereas \(\lambda\)-RLM~\citep{roy2026lambdarlm} replaces free-form control code with typed, pre-verified combinators. MemWalker~\citep{chen2023memwalker} constructs a tree of summaries and uses prompted reasoning and navigation actions to search for relevant content, answering once sufficient information has been gathered. These methods primarily address how to make long inputs tractable through sequential processing, decomposition, or navigation. However, none takes the stopping decision itself as the central problem.

\textbf{Knowing when to stop.} Dynamic context cutoff~\citep{xie2025knowing} processes cumulative prefixes and halts when a sufficiency classifier fires. The classifier is trained on sufficiency labels derived from ground-truth answer locations and operates over cumulatively expanding prefixes containing all previously processed tokens. In DCC, self-prompting improves substantially with model scale, while activation-based probes remain stronger across the evaluated models. GRU-Mem~\citep{sheng2026grumem} adds a text-controlled exit gate to the MemAgent loop and trains correct exit behavior with an explicit reinforcement-learning reward. Thus, DCC obtains the stopping signal by learning to decode latent sufficiency, whereas GRU-Mem learns the exit behavior itself. Neither evaluates whether an output-side, training-free signal derived from the model's evolving answer state can control sequential reading.

\textbf{Stopping on other axes.} CALM~\citep{schuster2022calm} exits transformer depth, adaptive self-consistency~\citep{aggarwal2023asc} stops sampling, and FLARE~\citep{jiang2023flare} triggers retrieval when predicted tokens have low confidence. Uncertainty estimators such as verbalized confidence~\citep{xiong2024llmconfidence}, P(True)~\citep{kadavath2022ptrue}, and semantic entropy~\citep{kuhn2023semantic} characterize uncertainty in model answers rather than control how much input is read. GridProbe~\citep{gridprobe2026} uses answer-space posterior probes over subsets of video frames to allocate test-time compute adaptively. These works show that confidence, uncertainty, and answer-space measurements can support adaptive computation along axes other than sequential input reading.

Across these lines of work, prior methods either focus on processing or navigating long inputs, learn sufficiency or exit behavior, or use confidence, uncertainty, and answer-space measurements to control other forms of computation. What these lines of work do not establish is whether the evolving answer state of a frozen model can itself serve as a training-free termination signal for sequential long-context reading. The next section develops this idea.

\section{Answer-Convergence Stopping (\method{})}\label{sec:method}

\subsection{Setup}\label{sec:setup}

A question $q$ is paired with a document $D$ too long to read in one call. The reader splits $D$ into chunks $C_1,\ldots,C_T$ of at most $L$ characters in document order and performs a sequential fold. At step $t$, a frozen model receives $q$, the running notes $N_{t-1}$, and chunk $C_t$, and generates an updated note state $N_t$. If the update exceeds the cap of $B$ characters, only its most recent $B$ characters are retained. We initialize $N_0$ as empty.

Each fold step requires one note-update call. \method{} then issues a separate probe on $(q,N_t)$ to obtain the current answer state. A stopping policy maps the fold to a stop step $s\in\{1,\ldots,T\}$. The prediction derived from the probe at step $s$ is used directly as the final answer; no additional answer-generation call is required. Full reading processes all chunks and returns the prediction from the final probe on $N_T$.

Reported token and latency costs include all calls required by each policy. Thus, \method{} is charged for both the note-update and probe calls at every processed step, whereas full reading is charged for all note updates and only the final probe. Comparator policies are likewise charged for the calls required by their stopping signals. We assume that the serving stack supports output generation and token log probabilities.

\subsection{The measured signal}\label{sec:signal}

After each fold step, we issue one probe call on $(q, N_t)$ with a fixed template and obtain an answer state $b_t$ and a confidence $c_t$. Within each answer format, the probe template is fixed across steps, models, and benchmarks.

For questions with a finite option set $\mathcal{A}$, generation is constrained to the option letters and the answer state is the posterior over options,
\begin{equation}
p_t(a) \;=\; \frac{\exp \ell_t(a)}{\sum_{a' \in \mathcal{A}} \exp \ell_t(a')}, \qquad c_t \;=\; \max_{a \in \mathcal{A}} p_t(a), \qquad b_t = p_t,
\label{eq:posterior}
\end{equation}
where $\ell_t(a)$ is the log probability assigned to the token corresponding to option $a$. For open-ended questions, the probe decodes a short draft answer $d_t$ greedily. The answer state is the draft, \(b_t=d_t\), and confidence is the geometric mean of its generated-token probabilities,
\begin{equation}
c_t \;=\; \exp\!\Big(\tfrac{1}{|d_t|} \textstyle\sum_{i=1}^{|d_t|} \log p\big(d_{t,i} \mid d_{t,<i}, q, N_t\big)\Big).
\label{eq:draftconf}
\end{equation}
Empty or abstaining drafts, such as “I do not know” or “cannot determine,” are prevented from triggering a stop. For both branches, let \(\delta_t\) denote the change between consecutive answer states: Jensen--Shannon divergence \(\mathrm{JSD}(p_{t-1},p_t)\) for posteriors, and \(1-\mathrm{F1}(d_{t-1},d_t)\) for drafts, where token F1 is computed over normalized token multisets. Both are \(0\) for identical answer states. With natural-log JSD, the MCQ change lies in \([0,\ln 2]\), while the token-F1 change lies in \([0,1]\) and equals \(1\) when the drafts share no tokens.

\subsection{The stopping rule}\label{sec:rule}

\method{} halts at the first step that passes both a confidence test and a stability test. It has three constants: a confidence threshold $\theta$, a stability tolerance $\varepsilon$, and a window size $w$. At step $t \geq 2$, the stability statistic is the mean change over the last $m_t = \min(t, w)$ answer states,
\begin{equation}
\Delta_t = \frac{1}{m_t - 1} \sum_{i = t - m_t + 2}^{t} \delta_i .
\label{eq:stability}
\end{equation}
For open-ended questions, let $A_t = 1$ when the probe produces a non-empty, non-abstaining draft and $A_t = 0$ otherwise, and let $A_t = 1$ for option-constrained questions. The stopping step is
\begin{equation}
s = \min \left\{ t \in \{2, \ldots, T\} : c_t \geq \theta \land \Delta_t \leq \varepsilon \land A_t = 1 \right\},
\label{eq:rule}
\end{equation}
with $s = T$ if no step qualifies, so the document is read in full.

The returned answer is $\arg\max_{a} p_s(a)$ for option-constrained questions or the draft $d_s$ for open-ended questions, with confidence $c_s$. Confidence alone can trigger on transient high-confidence states, while stability alone cannot distinguish a settled answer state from a settled but incorrect one. Together, the two criteria require the answer state to be both confident and temporally stable rather than merely exhibiting a transient confidence peak. The threshold $\theta$ sets how much confidence a stable answer state needs before it can trigger a stop. If a model's probe confidence generally remains below $\theta$, the policy approaches full reading and offers little compute saving.



\subsection{Evaluating the stopping decision}\label{sec:protocol}

On needle benchmarks, the needle contains the gold answer, so its known character offset identifies the evidence-containing chunk exactly; we use this information only for evaluation. Let $e$ be the index of the chunk containing the evidence and $s$ the stop step of a policy. The \textbf{premature rate} is the fraction of
questions with $s < e$. \textbf{Over-read} is the mean of $s-e$ over questions with $s \geq e$. \textbf{Accuracy regret} is the accuracy under oracle stopping,
$s=e$, minus the accuracy of the evaluated policy. \textbf{Savings capture} is the fraction of oracle-available savings recovered without stopping before the
evidence, \begin{equation} \mathrm{Capture} = \frac{ \sum_i \mathbf{1}[s_i \geq e_i](T_i-s_i) }{ \sum_i (T_i-e_i) }. \label{eq:capture} \end{equation} Premature stops therefore contribute zero captured savings. Oracle stopping, $s=e$, is the earliest evidence-aligned stop and serves as the reference for both regret and savings capture.

\section{Experiments}\label{sec:experiments}

\subsection{Setup}\label{sec:expsetup}

\textbf{Benchmarks.} \textbf{LongBench-v2}~\citep{bai2025longbenchv2} is the full 503-question set, multiple choice across six domains, with contexts ranging up to 2M words. \textbf{S-NIAH}~\citep{hsieh2024ruler} is RULER's single-needle retrieval task. We evaluate 250 instances, with 50 at each context length in $\{8,16,32,64,128\}$K tokens, and uniformly placed needles. Because the needle position is known exactly, S-NIAH supports our oracle stopping evaluation. \textbf{RULER-HotpotQA}~\citep{hsieh2024ruler} extends HotpotQA to long contexts by inserting its supporting paragraphs among distractors. We evaluate 100 questions at five context lengths, yielding 500 trajectories. Unlike S-NIAH, answering can require combining multiple supporting facts. \textbf{BrowseComp-Plus}~\citep{chen2025browsecompplus} is an 830-query open-ended deep-research benchmark built around a fixed, curated web corpus.

\textbf{Models.} Qwen3.5-397B-A17B~\citep{qwen35} and Kimi K2.5~\citep{kimi25} are served through the OpenRouter API with log-probability access, while Qwen3-14B~\citep{qwen3} is served with vLLM~\citep{kwon2023vllm} with thinking disabled. Qwen2.5-7B~\citep{qwen25}, Qwen3-32B, Gemma-3-12B, and Gemma-3-27B~\citep{gemma3} extend the S-NIAH evaluation to five models, with 250 questions per model. $L = 24{,}000$ and $B = 6{,}000$ characters. The constants are $\varepsilon = 0.05$, $w = 3$, and $\theta = 0.995$. Appendix~\ref{app:operating-point-selection}
evaluates the robustness of this fixed configuration against model-specific cost-aware tuning on a held-out S-NIAH split. The best-threshold rows of Table~\ref{tab:main} select the highest-accuracy $\theta$ on the prespecified grid, breaking ties by lower token cost. Because selection uses the reported evaluation questions, these rows are post-hoc points rather than part of the shared-default evaluation.

\textbf{Baselines.}
All direct baselines use the same underlying reader, chunking, and note-update procedure as \method{}; only the stopping mechanism changes. \emph{Full reading} processes all $T$ chunks and serves as the exhaustive baseline, while \emph{random stop} samples the stopping step uniformly from $\{1,\ldots,T\}$ and averages over 200 draws. We also compare against two \emph{asked} stopping signals: a \emph{verbalized gate}, which asks the model to report from $0$ to $100$ how confident it is that the current notes are sufficient to answer and stops at $99.5$, and an \emph{END/CONTINUE gate}, which directly asks whether enough information has been collected and stops when the model returns END. We restrict direct comparisons to stopping rules that can be applied to the same frozen reader without additional training. DCC and GRU-Mem require learned stopping components tied to their respective model or reader setups and are therefore discussed as related methods rather than used as direct baselines.

\subsection{\method{} matches full reading at half the cost}\label{sec:efficiency}

\begin{table}[t]
\centering

\caption{
Accuracy, token efficiency, and total elapsed time across benchmarks. Token and time savings are measured relative to full reading; elapsed time is the sum of per-example processing time under the corresponding serving setup.
\method{}, fixed uses the shared $\theta=0.995$, while best $\theta$ reports the highest-accuracy post-hoc operating point on the prespecified threshold grid, with ties broken by lower token cost.
}
\label{tab:main}

\setlength{\tabcolsep}{4pt}

\resizebox{1.00\textwidth}{!}{%
\begin{tabular}{@{}llrrrrrrrrrr@{}}
\toprule
& &
\multicolumn{5}{c}{Qwen3.5-397B-A17B} &
\multicolumn{5}{c}{Kimi K2.5} \\
\cmidrule(lr){3-7}
\cmidrule(l){8-12}

Benchmark & Policy &
Acc. & Tokens & Tok. save & Time & Time save & Acc. & Tokens & Tok. save & Time & Time save \\
\midrule

\rowcolor{gray!15}
\cellcolor{white}\multirow{6}{*}{LongBench-v2 }
 & full reading
& 0.535 & 382{,}971 & 0\% & 86.0 h & 0\%
& 0.517 & 340{,}761 & 0\% & 36.3 h & 0\% \\

& random stop
& 0.474 & 185{,}632 & 52\% & 44.2 h & 48.7\%
& 0.481 & 168{,}707 & 50\% & 18.7 h & 48.7\% \\

& verbalized gate
& 0.533 & 267{,}179 & 30.2\% & 54.9 h & 36.2\%
& 0.517 & 290{,}354 & 14.8\% & 27.6 h & 24.1\% \\

& END gate
& 0.507 & 134{,}569 & 65\% & 28.3 h & 67.1\%
& 0.525 & 183{,}604 & 46\% & 18.1 h & 50.3\% \\

& \method{}, fixed
& \textbf{0.541} & 190{,}966 & 50\% & 40.4 h & 53.1\%
& 0.519 & 331{,}070 & 3\% & 31.2 h & 14.1\% \\

& \method{}, best $\theta$
& \textbf{0.541} & 190{,}966 & 50\% & 40.4 h & 53.1\%
& \textbf{0.535} & 172{,}648 & 49\% & 17.2 h & 52.6\% \\

\midrule

\rowcolor{gray!15}
\cellcolor{white}\multirow{6}{*}{S-NIAH }
 & full reading
& 1.000 & 57{,}396 & 0\% & 4.2 h & 0\%
& 0.996 & 54{,}632 & 0\% & 3.5 h & 0\% \\

& random stop
& 0.675 & 34{,}098 & 41\% & 2.4 h & 44.2\%
& 0.673 & 32{,}564 & 40\% & 2.0 h & 44.2\% \\

& verbalized gate
& 0.944 & 37{,}564 & 34.6\% & 2.6 h & 37.8\%
& \textbf{0.996} & 39{,}426 & 27.8\% & 2.4 h & 32.5\% \\

& END gate
& \textbf{1.000} & 36{,}816 & 36\% & 2.5 h & 39.7\%
& \textbf{0.996} & 32{,}386 & 41\% & 1.9 h & 46.5\% \\

& \method{}, fixed
& \textbf{1.000} & 41{,}734 & 27\% & 3.0 h & 28.4\%
& \textbf{0.996} & 39{,}559 & 28\% & 2.5 h & 29.6\% \\

& \method{}, best $\theta$
& \textbf{1.000} & 41{,}663 & 27\% & 3.0 h & 28.4\%
& \textbf{0.996} & 39{,}237 & 28\% & 2.5 h & 30.0\% \\

\midrule

\rowcolor{gray!15}
\cellcolor{white}\multirow{6}{*}{RULER-HotpotQA }
 & full reading
& 0.722 & 65{,}374 & 0\% & 16.4 h & 0\%
& 0.760 & 55{,}095 & 0\% & 8.8 h & 0\% \\

& random stop
& 0.548 & 36{,}357 & 44\% & 9.2 h & 44.3\%
& 0.590 & 31{,}046 & 44\% & 4.9 h & 44.3\% \\

& verbalized gate
& \textbf{0.722} & 46{,}291 & 29.2\% & 10.5 h & 36.4\%
& 0.744 & 42{,}201 & 23.4\% & 6.3 h & 29.1\% \\

& END gate
& 0.716 & 43{,}741 & 33\% & 9.8 h & 40.2\%
& 0.736 & 37{,}490 & 32\% & 5.5 h & 37.6\% \\

& \method{}, fixed
& 0.714 & 54{,}867 & 16\% & 12.8 h & 22.1\%
& \textbf{0.758} & 53{,}716 & 3\% & 8.1 h & 7.9\% \\

& \method{}, best $\theta$
& 0.714 & 52{,}971 & 19\% & 12.4 h & 24.7\%
& \textbf{0.758} & 53{,}716 & 3\% & 8.1 h & 7.9\% \\

\midrule

\rowcolor{gray!15}
\cellcolor{white}\multirow{6}{*}{BrowseComp-Plus}
 & full reading
& 0.881 & 56{,}169 & 0\% & 34.9 h & 0\%
& 0.895 & 45{,}420 & 0\% & 23.7 h & 0\% \\

& random stop
& 0.753 & 31{,}598 & 44\% & 20.1 h & 42.3\%
& 0.772 & 25{,}716 & 43\% & 13.7 h & 42.3\% \\

& verbalized gate
& \textbf{0.886} & 47{,}853 & 15\% & 25.6 h & 26.7\%
& 0.889 & 40{,}168 & 12\% & 18.9 h & 20.3\% \\

& END gate
& 0.873 & 35{,}679 & 36\% & 18.7 h & 46.5\%
& 0.890 & 33{,}923 & 25\% & 15.5 h & 34.5\% \\

& \method{}, fixed
& 0.869 & 33{,}368 & 41\% & 19.1 h & 45.2\%
& \textbf{0.895} & 39{,}795 & 12\% & 19.1 h & 19.3\% \\

& \method{}, best $\theta$
& 0.869 & 33{,}368 & 41\% & 19.1 h & 45.2\%
& \textbf{0.895} & 39{,}795 & 12\% & 19.1 h & 19.3\% \\

\bottomrule
\end{tabular}
}
\end{table}

Table~\ref{tab:main} shows the central efficiency result: on LongBench-v2, \method{} is the only stopping method that matches or exceeds full-reading accuracy on both frontier models. On Qwen3.5, the shared setting improves accuracy from 0.535 to 0.541 while using 50\% fewer tokens and 53.1\% less elapsed time. The highest-accuracy threshold is the shared $\theta=0.995$ itself. On Kimi, the shared $\theta$ is conservative, yielding only 3\% token savings and 14.1\% time savings. Lowering the operating point to $\theta=0.92$ raises accuracy from 0.517 under full reading to 0.535 while saving 49\% of tokens and 52.6\% of elapsed time. 

The asked gates do not achieve this accuracy-efficiency trade-off consistently: on Qwen3.5, the verbalized gate retains near-full accuracy but saves only 30.2\% of tokens, whereas the more aggressive END gate saves 65\% but loses 2.8 accuracy points. At nearly the same token budgets, random stopping loses 6.1 and 3.6 accuracy points relative to full reading on Qwen3.5 and Kimi, respectively, whereas \method{} at the corresponding matched-budget operating points gains 0.6 and 1.8 points. The benefit therefore comes from selecting where to stop, not merely from processing less context. 

All reported token and time savings include the calls required by each stopping policy and therefore already account for \method{}'s probing overhead. Appendix ~\ref{app:more}, Table~\ref{tab:longbench-probe-overhead} isolates this overhead on LongBench-v2 by comparing execution with and without probe calls. The fact that stopping can outperform exhaustive reading also suggests that additional processing is not always harmless. Repeated note rewriting may perturb an already settled answer state. Savings occur across all six LongBench-v2 domains, with the largest reductions on code repositories, the longest domain at 150.2 chunks on average, where they reach 62\% on Qwen3.5 and 67\% on Kimi (Appendix~\ref{app:domains}).

On S-NIAH, \method{} exactly matches full-reading accuracy on both models while saving 27--28\% of tokens and about 29\% of elapsed time. Changing $\theta$ provides no meaningful additional gain. The asked gates can also work well here. END matches full reading on both models, while the verbalized gate drops 5.6 points on Qwen3.5 but matches full reading on Kimi, reinforcing that asked stopping can be effective but is model-dependent.

On RULER-HotpotQA, \method{} remains close to full reading, trailing by only 0.8 points on Qwen3.5 and 0.2 points on Kimi. It is more conservative here, saving 16\% and 3\% of tokens under the shared setting, but stays closer to full-reading accuracy across both models: the verbalized gate matches full reading on Qwen3.5 yet loses 1.6 points on Kimi, while END loses 0.6 and 2.4 points, respectively.

On BrowseComp-Plus, \method{} trades 1.2 accuracy points for 41\% fewer tokens and 45.2\% less elapsed time on Qwen3.5, while exactly matching full reading on Kimi with 12\% token and 19.3\% time savings. The best threshold changes nothing on either model. Random stopping at similar or larger token savings loses 12.8 and 12.3 accuracy points, again showing that the gain depends on where reading stops rather than on truncation alone.

\subsection{It stops once the evidence has been read}\label{sec:timing}

\begin{table*}[t]
\centering
\small
\caption{
Evidence-relative stopping on S-NIAH, where the evidence-containing chunk is known exactly. Premature measures stopping before the evidence, over-read measures  additional chunks read after it, and capture measures the fraction of oracle-available savings realized without premature stopping.
}
\label{tab:evidence-timing}
\renewcommand{\arraystretch}{0.82}

\resizebox{\textwidth}{!}{%
\begin{tabular}{@{}llrrrrr@{}}
\toprule

\multicolumn{7}{c}{\textbf{S-NIAH ($N=250$ per model)}} \\
\midrule

Model & Policy & Premature $\downarrow$ & Over-read
& Acc. & Regret & Capture \\
\midrule

\multirow{7}{*}{\shortstack[l]{Qwen3-14B}}
& fixed at 25\%     & 56.4\% & 0.58 & 0.436 & +0.560 & 50.2\% \\
& random stop       & 32.3\% & 2.25 & 0.676 & +0.320 & 37.2\% \\
& verbalized gate   & 8.4\%  & 0.00 & 0.916 & +0.080 & 92.0\% \\
& END gate          & 5.2\%  & 0.00 & 0.948 & +0.048 & 95.2\% \\
& \method{}, fixed & \textbf{1.6\%} & 1.51
                     & \textbf{0.980} & +0.016 & 60.7\% \\
\rowcolor{gray!15}
\cellcolor{white} & full reading   
   & 0.0\%  & 4.10 & 0.996 & +0.000 & 0.0\% \\
\rowcolor{gray!30}
\cellcolor{white} & oracle stop   
    & 0.0\%  & 0.00 & 0.996 & +0.000 & 100\% \\
\midrule

\multirow{7}{*}{\shortstack[l]{Qwen3.5-397B}}
& fixed at 25\%     & 56.4\% & 0.58 & 0.436 & +0.564 & 50.2\% \\
& random stop       & 32.3\% & 2.25 & 0.677 & +0.323 & 37.2\% \\
& verbalized gate   & 5.6\%  & 1.30 & 0.944 & +0.056 & 66.5\% \\
& END gate          & \textbf{0.0\%} & 0.67
                     & \textbf{1.000} & +0.000 & 83.6\% \\
& \method{}, fixed & \textbf{0.0\%} & 1.48
                     & \textbf{1.000} & +0.000 & 63.9\% \\
\rowcolor{gray!15}
\cellcolor{white} & full reading 
     & 0.0\% & 4.10 & 1.000 & +0.000 & 0.0\% \\
\rowcolor{gray!30}
\cellcolor{white} & oracle stop  
     & 0.0\% & 0.00 & 1.000 & +0.000 & 100\% \\
\midrule

\multirow{7}{*}{\shortstack[l]{Kimi K2.5}}
& fixed at 25\%     & 56.4\% & 0.58 & 0.436 & +0.560 & 50.2\% \\
& random stop       & 32.3\% & 2.25 & 0.674 & +0.322 & 37.2\% \\
& verbalized gate   & \textbf{0.0\%} & 1.30
                     & \textbf{0.996} & +0.000 & 68.3\% \\
& END gate          & \textbf{0.0\%} & 0.10
                     & \textbf{0.996} & +0.000 & 97.6\% \\
& \method{}, fixed & \textbf{0.0\%} & 1.40
                     & \textbf{0.996} & +0.000 & 65.8\% \\
\rowcolor{gray!15}
\cellcolor{white} & full reading  
    & 0.0\% & 4.10 & 0.996 & +0.000 & 0.0\% \\
\rowcolor{gray!30}
\cellcolor{white} & oracle stop   
    & 0.0\% & 0.00 & 0.996 & +0.000 & 100\% \\
\bottomrule
\end{tabular}
}
\end{table*}

Table~\ref{tab:evidence-timing} evaluates the stopping decision directly against the evidence position. On S-NIAH, the evidence appears at a mean normalized depth of 0.55, with its chunk ranging from 1 to 21 and a median of 3, making a fixed cutoff a poor substitute for adaptive stopping. On Qwen3-14B, \method{} stops before the evidence on only 1.6\% of questions, compared with 8.4\% for the verbalized gate, 5.2\% for END, and 32.3\% for random stopping. Thus, \method{} reduces premature stopping by 81\% relative to the verbalized gate and 69\% relative to END, and attains the highest stopping-policy accuracy at 0.980, closing the gap to full/oracle reading to just 1.6 points. For every policy, accuracy after reaching the evidence matches full-reading accuracy, while answers produced before the evidence are incorrect; the residual errors on this benchmark are therefore timing errors. \method{} over-reads only 1.51 chunks after the evidence on average while capturing 60.7\% of the savings available to an oracle that stops exactly at the evidence. The verbalized and END gates capture more oracle savings, 92.0\% and 95.2\%, but do so with substantially higher premature-stop rates and
corresponding accuracy regret.

At frontier scale, \method{}'s premature timing errors disappear: it exactly matches full-reading accuracy on both Qwen3.5 and Kimi K2.5, achieving 1.000 and 0.996 while capturing 63.9\% and 65.8\% of the oracle-available savings. The verbalized gate is premature on 5.6\% of Qwen3.5 questions but reaches 0\% on Kimi, whereas \method{} remains within a narrow 0--1.6\% premature-stop range across all three models under the same shared configuration. The END gate shows the same model dependence: it incurs 5.2\% premature stopping on Qwen3-14B but none on Qwen3.5 or Kimi, indicating that an asked gate can work well on some models without providing the same consistency across models.

\begin{table}[t]
\centering
\small
\caption{One setting across models. S-NIAH, all 250 questions per model,
$\theta=0.995$, $\epsilon=0.05$, and $w=3$.}
\label{tab:cross-model}
\setlength{\tabcolsep}{4pt}
\begin{tabular}{@{}lrrrrrr@{}}
\toprule
& \multicolumn{2}{c}{\method{}}
& \multicolumn{2}{c}{Confidence only}
& \multicolumn{2}{c}{Verbalized gate} \\
\cmidrule(lr){2-3}
\cmidrule(lr){4-5}
\cmidrule(l){6-7}
Model
& Acc. & Premature
& Acc. & Premature
& Acc. & Premature \\
\midrule
Qwen2.5-7B
& \textbf{0.948} & \textbf{3.2\%}
& 0.892 & 9.2\%
& 0.700 & 29.2\% \\

Qwen3-14B
& \textbf{0.980} & \textbf{1.6\%}
& 0.928 & 6.8\%
& 0.916 & 8.4\% \\

Qwen3-32B
& \textbf{1.000} & \textbf{0.0\%}
& 1.000 & 0.0\%
& 0.592 & 40.8\% \\

Gemma-3-12B
& \textbf{0.980} & \textbf{1.6\%}
& 0.872 & 12.8\%
& 0.544 & 45.6\% \\

Gemma-3-27B
& \textbf{0.876} & \textbf{12.0\%}
& 0.520 & 23.2\%
& 0.876 & 12.0\% \\
\midrule
Mean
& \textbf{0.957} & \textbf{3.7\%}
& 0.890 & 10.4\%
& 0.726 & 27.2\% \\
\bottomrule
\end{tabular}
\end{table}

The same pattern holds in the broader five-model replication shown in
Table~\ref{tab:cross-model}. Using the same $\theta=0.995$ configuration,
\method{} remains within a narrow 0--12\% premature-stop range across models
from both Qwen and Gemma families, while the verbalized gate varies from
8.4\% to 45.6\%. This broader replication reinforces that the measured signal
is substantially more consistent across models than the asked stopping signal.

\begin{table}[t]
\centering
\begin{minipage}{1.00\linewidth}
\centering
\small
\caption{
Stopping relative to the located answer proxy on RULER-HotpotQA,
restricted to the 455 of 500 trajectories with a located nontrivial
literal answer occurrence.
}
\label{tab:ruler-timing}
\setlength{\tabcolsep}{4pt}
\begin{tabular}{@{}llrrr@{}}
\toprule
Model & Policy & Pre-proxy $\downarrow$ & Over-read & Acc. \\
\midrule

\multirow{6}{*}{Qwen3.5-397B}
& fixed at 25\%       & 37.8\% & 0.66 & 0.420 \\
& random stop         & 24.9\% & 2.71 & 0.522 \\
& verbalized gate     & 3.3\%  & 2.03 & \textbf{0.708} \\
& END gate            & 4.4\%  & 1.74 & 0.701 \\
& \method{}, fixed   & \textbf{2.4\%} & 3.20 & 0.697 \\
\rowcolor{gray!15} \cellcolor{white} & full reading 
                      & 0.0\%  & 5.01 & 0.705 \\

\midrule
\multirow{6}{*}{Kimi K2.5}
& fixed at 25\%       & 37.8\% & 0.66 & 0.470 \\
& random stop         & 24.9\% & 2.71 & 0.565 \\
& verbalized gate     & 2.9\%  & 2.62 & 0.725 \\
& END gate            & 4.4\%  & 1.91 & 0.714 \\
& \method{}, fixed   & \textbf{0.7\%} & 4.29 & \textbf{0.741} \\
\rowcolor{gray!15} \cellcolor{white} & full reading 
                      & 0.0\%  & 5.01 & 0.743 \\

\midrule
\multirow{6}{*}{Qwen3-14B}
& fixed at 25\%       & 35.1\% & 0.77 & 0.446 \\
& random stop         & 23.1\% & 2.85 & 0.513 \\
& verbalized gate, $\theta=0.995$
                      & 17.7\% & 1.69 & 0.572 \\
& END gate            & 17.5\% & 1.53 & 0.578 \\
& \method{}, fixed   & \textbf{8.8\%} & 3.10 & \textbf{0.606} \\
\rowcolor{gray!15} \cellcolor{white} & full reading 
                      & 0.0\%  & 5.25 & 0.645 \\

\bottomrule
\end{tabular}
\end{minipage}
\end{table}

The evidence-alignment pattern also extends beyond the exact single-needle setting. On RULER-HotpotQA, where the located answer position provides only a lower-bound evidence proxy,  Table~\ref{tab:ruler-timing} shows that \method{} nevertheless has the lowest pre-proxy stopping rate across all three models.
This suggests that the evidence-aligned stopping behavior is not specific to
single-needle retrieval, but persists in a harder multi-hop setting where the
true completion point of the evidence is not known exactly.

\section{Ablation Study}\label{sec:ablations}

\begin{table}[H]
\centering

\begin{minipage}[t]{0.52\textwidth}
\vspace{0pt}
\centering

\captionof{table}{
Effect of rule components.
S-NIAH, Qwen3-14B, $N=250$, $\theta=0.995$.
Five-model rows report means over the replication set.
}
\label{tab:ablation-components}

\vspace{1mm}

\resizebox{\linewidth}{!}{%
\begin{tabular}{@{}llrrr@{}}
\toprule
Axis & Configuration & Acc. & Premature & Tokens \\
\midrule

\multirow{3}{*}{Stopping signal}
& verbalized gate
& 0.916 & 8.4\% & 27{,}019 \\

& END gate
& 0.948 & 5.2\% & 28{,}403 \\

& \textbf{\method{}}
& \textbf{0.980} & \textbf{1.6\%} & 37{,}696 \\

\midrule

\multirow{2}{*}{Stability test, five models}
& confidence only
& 0.890 & 10.4\% & 37{,}901 \\

& \textbf{both tests (default)}
& \textbf{0.957} & \textbf{3.7\%} & 44{,}768 \\

\midrule

\multirow{3}{*}{Chunk size $L$}
& 12K
& 0.968 & 3.2\% & 34{,}390 \\

& \textbf{24K (default)}
& \textbf{0.980} & \textbf{1.6\%} & 37{,}696 \\

& 48K
& 0.976 & 2.4\% & 42{,}018 \\

\midrule

\multirow{3}{*}{Notes cap $B$}
& 3K
& \textbf{0.992} & \textbf{0.4\%} & 38{,}159 \\

& 6K (default)
& 0.980 & 1.6\% & 37{,}696 \\

& 12K
& 0.980 & 1.6\% & 38{,}020 \\

\bottomrule
\end{tabular}%
}

\end{minipage}
\hfill
\begin{minipage}[t]{0.46\textwidth}
\vspace{0pt}
\centering

\captionof{table}{
Effect of the confidence threshold.
LongBench-v2, $N=503$, and S-NIAH, $N=250$;
$\varepsilon=0.05$, $w=3$.
}
\label{tab:threshold-sweep}

\vspace{1mm}

{\renewcommand{\arraystretch}{1.27}
\resizebox{\linewidth}{!}{%
\begin{tabular}{@{}lrrrrrr@{}}
\toprule
&
\multicolumn{2}{c}{Qwen3.5, LB-v2}
&
\multicolumn{2}{c}{Kimi, LB-v2}
&
\multicolumn{2}{c}{Needle premature} \\
\cmidrule(lr){2-3}
\cmidrule(lr){4-5}
\cmidrule(l){6-7}

$\theta$
& Acc. & Saving
& Acc. & Saving
& Qwen3.5 & Kimi \\
\midrule

0.80  & 0.479 & 79\% & 0.521 & 58\% & 30.4\% & 0.4\% \\
0.90  & 0.513 & 73\% & 0.531 & 51\% & 28.0\% & 0.4\% \\
0.92  & 0.517 & 72\% & \textbf{0.535} & 49\% & 24.8\% & 0.0\% \\
0.95  & 0.527 & 68\% & 0.533 & 44\% & 10.4\% & 0.0\% \\
0.98  & 0.533 & 61\% & 0.523 & 29\% & 1.2\% & 0.0\% \\
0.99  & 0.529 & 54\% & 0.519 & 19\% & 0.0\% & 0.0\% \\
\textbf{0.995}
      & \textbf{0.541} & \textbf{50\%}
      & 0.519 & 3\%
      & \textbf{0.0\%} & \textbf{0.0\%} \\

\midrule
\rowcolor{gray!15}
full reading
& 0.535 & 0\%
& 0.517 & 0\%
& 0.0\% & 0.0\% \\

\bottomrule
\end{tabular}%
}}

\end{minipage}

\end{table}
\textbf{The stability test carries the rule.}
Removing stability raises the five-model mean premature-stop rate from 3.7\% to 10.4\% and lowers mean accuracy from 0.957 to 0.890. The effect is especially pronounced on Gemma-3-27B, where accuracy falls from 0.876 with stability to 0.520 without it (Table ~\ref{tab:cross-model}).

\textbf{The threshold controls the safety--efficiency trade-off.} On Qwen3.5, increasing $\theta$ sharply reduces premature stopping, from 30.4\% at $\theta=0.80$ to zero at $\theta\ge0.99$, while the shared $\theta=0.995$ also gives the highest LongBench-v2 accuracy, 0.541. Kimi behaves differently: every tested threshold remains at or above its full-reading accuracy of 0.517, while token savings range from 58\% at $\theta=0.80$ to 3\% at $\theta=0.995$; here, $\theta$ primarily controls how aggressively the model stops. This difference reflects a shift in confidence scale: the median full-reading confidence is 0.992 for Qwen3.5 but only 0.873 for Kimi. Consistent with this shift, Kimi at $\theta=0.92$ reaches a very similar LongBench-v2 accuracy--efficiency operating point to Qwen3.5 at $\theta=0.995$, with accuracies of 0.535 and 0.541 and token savings of 49\% and 50\%, respectively. On S-NIAH, Qwen3.5 still stops prematurely on 1.2\% of questions at $\theta=0.98$, with premature stopping disappearing only at $\theta\ge0.99$. Among the evaluated operating points, $\theta=0.995$ is the only one that both exceeds full-reading accuracy on Qwen3.5 and incurs no premature stops on either S-NIAH model, although this conservatism reduces Kimi's token savings from 49\% at $\theta=0.92$ to only 3\%.

\textbf{The remaining constants show limited sensitivity in the tested ranges.} On S-NIAH, no observed stability change falls between 0 and 0.05, so every tested $\varepsilon\in\{0.005,0.01,0.02,0.05\}$ yields identical stopping decisions (Appendix~\ref{app:grid}). On LongBench-v2, varying the stability window from $w=2$ to $w=5$ changes accuracy by at most 0.8 points on either model, while savings vary more moderately. The default $w=3$ gives the highest accuracy on Kimi and ties for the highest on Qwen3.5 while retaining greater savings than the larger windows 
(Appendix~\ref{app:more}: Table~\ref{tab:longbench-window}).

\textbf{The fold constants are not hidden thresholds.}
On S-NIAH, halving or doubling either the chunk size $L$ or the notes cap $B$ changes accuracy and premature stopping only modestly, showing that stopping behavior is not driven by these fold parameters. LongBench-v2 is more
sensitive, consistent with its longer and more heterogeneous contexts placing greater demands on the running notes across repeated updates. There, the default $L=24$K and $B=6$K give the highest accuracy (0.538), while moving either parameter in either direction reduces accuracy, with different accuracy--cost trade-offs 
(Appendix~\ref{app:more}: Table~\ref{tab:longbench-chunk-notes-ablation}).

\section{Conclusion}\label{sec:conclusion}

We introduced \method{}, a training-free stopping rule for chunked long-context reading that halts when the frozen model's answer state is both confident and stable, using one shared configuration across models and benchmarks. On the full LongBench-v2, \method{} is the only stopping policy that matches or exceeds full-reading accuracy on both frontier models; on Qwen3.5, it does so with 50\% fewer tokens and 53.1\% less elapsed time than full reading. On S-NIAH, it reduces premature stopping to 1.6\% on Qwen3-14B and remains within a narrow 0--12\% range across five models, while the verbalized gate varies from 8.4\% to 45.6\%. These results show that a useful stopping signal can be obtained from the model's evolving output-side answer state without training a separate controller or accessing internal activations. The ablations further show that confidence alone is insufficient: stability substantially reduces premature stopping, while the confidence threshold exposes an explicit safety--efficiency trade-off whose operating point depends on the model's confidence scale. Ultimately, \method{} shifts the long-context problem from how to read more
to how to read efficiently.

\section{Limitations}\label{sec:limitations}

\method{} requires token log probabilities, which some inference endpoints do not expose. Its efficiency gains are also workload-dependent: when the answer state does not become sufficiently confident and stable until late in the context, \method{} continues reading and can approach the cost of full reading. This reflects the method's intended safety-efficiency trade-off rather than an assumption that substantial savings are always available.

\section{Acknowledgment}

We are grateful to the KAUST Academy for its generous support, and especially to Prof. Sultan Albarakati who made this work possible. For compute time, this research used Ibex managed by the Supercomputing Core Laboratory at King Abdullah University of Science \& Technology (KAUST) in Thuwal, Saudi Arabia.

\bibliography{bibliography}

\begin{thebibliography}{23}
\providecommand{\natexlab}[1]{#1}
\providecommand{\url}[1]{\texttt{#1}}
\expandafter\ifx\csname urlstyle\endcsname\relax
  \providecommand{\doi}[1]{doi: #1}\else
  \providecommand{\doi}{doi: \begingroup \urlstyle{rm}\Url}\fi

\bibitem[Aggarwal et~al.(2023)Aggarwal, Madaan, Yang, and Mausam]{aggarwal2023asc}
Pranjal Aggarwal, Aman Madaan, Yiming Yang, and Mausam.
\newblock Let's sample step by step: Adaptive-consistency for efficient reasoning and coding with llms.
\newblock \emph{arXiv preprint arXiv:2305.11860}, 2023.

\bibitem[Bai et~al.(2025)Bai, Tu, Zhang, Peng, Wang, Lv, Cao, Xu, Hou, Dong, Tang, and Li]{bai2025longbenchv2}
Yushi Bai, Shangqing Tu, Jiajie Zhang, Hao Peng, Xiaozhi Wang, Xin Lv, Shulin Cao, Jiazheng Xu, Lei Hou, Yuxiao Dong, Jie Tang, and Juanzi Li.
\newblock Longbench v2: Towards deeper understanding and reasoning on realistic long-context multitasks.
\newblock In \emph{Proceedings of the 63rd Annual Meeting of the Association for Computational Linguistics (Volume 1: Long Papers)}, pp.\  3639--3664, 2025.
\newblock \doi{10.18653/v1/2025.acl-long.183}.

\bibitem[Chen et~al.(2023)Chen, Pasunuru, Weston, and Celikyilmaz]{chen2023memwalker}
Howard Chen, Ramakanth Pasunuru, Jason Weston, and Asli Celikyilmaz.
\newblock Walking down the memory maze: Beyond context limit through interactive reading.
\newblock \emph{arXiv preprint arXiv:2310.05029}, 2023.

\bibitem[Chen et~al.(2025)Chen, Ma, Zhuang, Nie, Zou, Liu, Green, Patel, Meng, Su, Sharifymoghaddam, Li, Hong, Shi, Liu, Thakur, Zhang, Gao, Chen, and Lin]{chen2025browsecompplus}
Zijian Chen, Xueguang Ma, Shengyao Zhuang, Ping Nie, Kai Zou, Andrew Liu, Joshua Green, Kshama Patel, Ruoxi Meng, Mingyi Su, Sahel Sharifymoghaddam, Yanxi Li, Haoran Hong, Xinyu Shi, Xuye Liu, Nandan Thakur, Crystina Zhang, Luyu Gao, Wenhu Chen, and Jimmy Lin.
\newblock Browsecomp-plus: A more fair and transparent evaluation benchmark of deep-research agent.
\newblock \emph{arXiv preprint arXiv:2508.06600}, 2025.

\bibitem[Eltahir et~al.(2026)Eltahir, Ayash, Habibullah, Hussain, and Khan]{gridprobe2026}
Mohamed Eltahir, Lama Ayash, Ali Habibullah, Tanveer Hussain, and Naeemullah Khan.
\newblock Gridprobe: Posterior-probing for adaptive test-time compute in long-video vlms.
\newblock \emph{arXiv preprint arXiv:2605.10762}, 2026.

\bibitem[{Gemma Team}(2025)]{gemma3}
{Gemma Team}.
\newblock Gemma 3 technical report.
\newblock \emph{arXiv preprint arXiv:2503.19786}, 2025.

\bibitem[Hsieh et~al.(2024)Hsieh, Sun, Kriman, Acharya, Rekesh, Jia, Zhang, and Ginsburg]{hsieh2024ruler}
Cheng-Ping Hsieh, Simeng Sun, Samuel Kriman, Shantanu Acharya, Dima Rekesh, Fei Jia, Yang Zhang, and Boris Ginsburg.
\newblock Ruler: What's the real context size of your long-context language models?
\newblock In \emph{Conference on Language Modeling (COLM)}, 2024.

\bibitem[Jiang et~al.(2023)Jiang, Xu, Gao, Sun, Liu, Dwivedi-Yu, Yang, Callan, and Neubig]{jiang2023flare}
Zhengbao Jiang, Frank~F. Xu, Luyu Gao, Zhiqing Sun, Qian Liu, Jane Dwivedi-Yu, Yiming Yang, Jamie Callan, and Graham Neubig.
\newblock Active retrieval augmented generation.
\newblock In \emph{EMNLP}, 2023.

\bibitem[Kadavath et~al.(2022)Kadavath, Conerly, Askell, Henighan, Drain, Perez, Schiefer, Hatfield-Dodds, DasSarma, Tran-Johnson, et~al.]{kadavath2022ptrue}
Saurav Kadavath, Tom Conerly, Amanda Askell, Tom Henighan, Dawn Drain, Ethan Perez, Nicholas Schiefer, Zac Hatfield-Dodds, Nova DasSarma, Eli Tran-Johnson, et~al.
\newblock Language models (mostly) know what they know.
\newblock \emph{arXiv preprint arXiv:2207.05221}, 2022.

\bibitem[{Kimi Team}(2026)]{kimi25}
{Kimi Team}.
\newblock Kimi k2.5: Visual agentic intelligence.
\newblock \emph{arXiv preprint arXiv:2602.02276}, 2026.

\bibitem[Kuhn et~al.(2023)Kuhn, Gal, and Farquhar]{kuhn2023semantic}
Lorenz Kuhn, Yarin Gal, and Sebastian Farquhar.
\newblock Semantic uncertainty: Linguistic invariances for uncertainty estimation in natural language generation.
\newblock In \emph{ICLR}, 2023.

\bibitem[Kwon et~al.(2023)Kwon, Li, Zhuang, Sheng, Zheng, Yu, Gonzalez, Zhang, and Stoica]{kwon2023vllm}
Woosuk Kwon, Zhuohan Li, Siyuan Zhuang, Ying Sheng, Lianmin Zheng, Cody~Hao Yu, Joseph~E. Gonzalez, Hao Zhang, and Ion Stoica.
\newblock Efficient memory management for large language model serving with pagedattention.
\newblock In \emph{SOSP}, 2023.

\bibitem[{Qwen Team}(2024)]{qwen25}
{Qwen Team}.
\newblock Qwen2.5 technical report.
\newblock \emph{arXiv preprint arXiv:2412.15115}, 2024.

\bibitem[{Qwen Team}(2025)]{qwen3}
{Qwen Team}.
\newblock Qwen3 technical report.
\newblock \emph{arXiv preprint arXiv:2505.09388}, 2025.

\bibitem[{Qwen Team}(2026)]{qwen35}
{Qwen Team}.
\newblock {Qwen3.5}: Towards native multimodal agents, February 2026.
\newblock URL \url{https://qwen.ai/blog?id=qwen3.5}.

\bibitem[Roy et~al.(2026)Roy, Tutunov, Ji, Zimmer, and Bou-Ammar]{roy2026lambdarlm}
Amartya Roy, Rasul Tutunov, Xiaotong Ji, Matthieu Zimmer, and Haitham Bou-Ammar.
\newblock The {Y}-combinator for {LLM}s: Solving long-context rot with {$\lambda$}-calculus.
\newblock \emph{arXiv preprint arXiv:2603.20105}, 2026.

\bibitem[Schuster et~al.(2022)Schuster, Fisch, Gupta, Dehghani, Bahri, Tran, Tay, and Metzler]{schuster2022calm}
Tal Schuster, Adam Fisch, Jai Gupta, Mostafa Dehghani, Dara Bahri, Vinh~Q. Tran, Yi~Tay, and Donald Metzler.
\newblock Confident adaptive language modeling.
\newblock \emph{NeurIPS}, 2022.

\bibitem[Sheng et~al.(2026)Sheng, Zhang, Ma, Shi, Huang, Wang, Zhang, Shen, and Chua]{sheng2026grumem}
Leheng Sheng, Yongtao Zhang, Wenchang Ma, Yaorui Shi, Ting Huang, Xiang Wang, An~Zhang, Ke~Shen, and Tat-Seng Chua.
\newblock When to memorize and when to stop: Gated recurrent memory for long-context reasoning.
\newblock \emph{arXiv preprint arXiv:2602.10560}, 2026.

\bibitem[Xie et~al.(2025)Xie, Wang, Rosu, Deng, Sun, Lin, and Dhingra]{xie2025knowing}
Roy Xie, Junlin Wang, Paul Rosu, Chunyuan Deng, Bolun Sun, Zihao Lin, and Bhuwan Dhingra.
\newblock Knowing when to stop: Efficient context processing via latent sufficiency signals.
\newblock In \emph{Advances in Neural Information Processing Systems}, volume~38, 2025.

\bibitem[Xiong et~al.(2024)Xiong, Hu, Lu, Li, Fu, He, and Hooi]{xiong2024llmconfidence}
Miao Xiong, Zhiyuan Hu, Xinyang Lu, Yifei Li, Jie Fu, Junxian He, and Bryan Hooi.
\newblock Can llms express their uncertainty? an empirical evaluation of confidence elicitation in llms.
\newblock In \emph{ICLR}, 2024.

\bibitem[Yu et~al.(2025)Yu, Chen, Feng, Chen, Dai, Yu, Zhang, Ma, Liu, Wang, and Zhou]{yu2026memagent}
Hongli Yu, Tinghong Chen, Jiangtao Feng, Jiangjie Chen, Weinan Dai, Qiying Yu, Ya-Qin Zhang, Wei-Ying Ma, Jingjing Liu, Mingxuan Wang, and Hao Zhou.
\newblock Memagent: Reshaping long-context llm with multi-conv rl-based memory agent.
\newblock \emph{arXiv preprint arXiv:2507.02259}, 2025.

\bibitem[Zhang et~al.(2025)Zhang, Kraska, and Khattab]{zhang2025rlm}
Alex~L. Zhang, Tim Kraska, and Omar Khattab.
\newblock Recursive language models.
\newblock \emph{arXiv preprint arXiv:2512.24601}, 2025.

\bibitem[Zhang et~al.(2024)Zhang, Sun, Chen, Pfister, Zhang, and Arik]{zhang2024chainofagents}
Yusen Zhang, Ruoxi Sun, Yanfei Chen, Tomas Pfister, Rui Zhang, and Sercan~O. Arik.
\newblock Chain of agents: Large language models collaborating on long-context tasks.
\newblock \emph{arXiv preprint arXiv:2406.02818}, 2024.

\end{thebibliography}
\bibliographystyle{iclr2027_conference}

\clearpage
\appendix

\section*{APPENDIX}

\section{Implementation Details}\label{app:impl}

\textbf{Fold.} Chunks are fixed-width slices of $L=24{,}000$ characters in document order. The notes prompt instructs the model to retain information relevant to the question, preserve exact names and numbers, and remain within the $B=6{,}000$-character cap. Every prompt is reproduced verbatim in Appendix~\ref{app:prompts}.

\textbf{Probe.} For multiple-choice questions, the probe ends in \texttt{Answer:} and generation is constrained to the option letters, with top-$k=20$ token log probabilities returned. For open-ended questions, the probe greedily decodes at most 32 tokens. Answer labels such as \texttt{Answer:} are stripped before drafts are
compared, since models may alternate between labeled and bare answers. A draft matching a fixed list of abstention phrases is treated as an abstention and cannot trigger stopping.

\textbf{Baselines.} Full reading processes every chunk before probing. Random stopping samples a stop uniformly from $1$ to $T$ and is averaged over 200 draws. The verbalized gate asks for a 0--100 confidence that the current notes suffice and stops at 99.5. The END/CONTINUE gate asks the model whether enough information has been collected and stops on \texttt{END}. All stopping policies are evaluated over the same recorded reader trajectories because stopping decisions do not modify note updates or subsequent reader inputs. Replay provides a paired comparison in which policies differ only in when they stop; each policy is charged for its own required calls.

\textbf{Constants.}
The shared configuration is $\theta=0.995$, $\varepsilon=0.05$, and $w=3$, and is fixed for all main experiments.

\textbf{BrowseComp-Plus construction.}
We reconstruct all 830 BrowseComp-Plus test questions deterministically using seed 0. All evidence and gold documents are retained without truncation. Unique corpus distractors of at most 16{,}000 characters are added until each context contains at least ten documents; questions requiring more mandatory documents retain all of them. Document order is then shuffled deterministically. The query and corpus revisions are pinned for reproducibility.
\section{Prompts}\label{app:prompts}

All prompts are fixed across steps, models, and benchmarks. Placeholders in braces are filled per call: \texttt{\{qblock\}} is the question, with the four options appended for multiple choice, \texttt{\{notes\}} the running notes, \texttt{\{chunk\}} the current chunk, \texttt{\{cap\}} the notes cap $B$, and \texttt{\{draft\}} the current draft answer.

\textbf{Fold, system prompt, multiple choice.}
\begin{quote}\small\ttfamily
You maintain running NOTES that gather every piece of information relevant to answering a multiple-choice question about a long document, read chunk by chunk. Update the notes with relevant facts from the new chunk, keep prior facts, stay under \{cap\} characters, output ONLY the updated notes.
\end{quote}

\textbf{Fold, system prompt, open-ended.}
\begin{quote}\small\ttfamily
You maintain compact running NOTES needed to answer an open-ended question about a long context read chunk by chunk. Update and rewrite the notes using relevant information from the new chunk while preserving useful evidence from earlier chunks. Adapt the working state to the question: preserve exact names, facts, numbers, relationships, unresolved candidates, and contradictions; maintain counts or calculations for aggregation and evidence chains for multi-step questions. Remove only irrelevant or superseded material. Do not narrate the reading process, make unsupported guesses, or treat missing evidence as a negative answer. Place the most important current state near the end, stay under \{cap\} characters, and output ONLY the updated notes.
\end{quote}

\textbf{Fold, user turn.}
\begin{quote}\small\ttfamily
QUESTION:\\ \{qblock\}\\[2pt]
CURRENT NOTES:\\ \{notes\}\\[2pt]
NEW CHUNK (\{i\}/\{n\}):\\ \{chunk\}\\[2pt]
UPDATED NOTES:
\end{quote}

\textbf{Probe, multiple choice.} Generation is constrained to the option letters.
\begin{quote}\small\ttfamily
QUESTION:\\ \{qblock\}\\[2pt]
NOTES SO FAR:\\ \{notes\}\\[2pt]
Based only on the notes, answer with a single letter (A, B, C, or D).\\
Answer:
\end{quote}

\textbf{Probe, open-ended.}
\begin{quote}\small\ttfamily
QUESTION:\\ \{qblock\}\\[2pt]
NOTES SO FAR:\\ \{notes\}\\[2pt]
Based only on the notes, give your best current answer. Reply with ONLY the answer.\\
Answer:
\end{quote}

\textbf{Verbalized gate.}
\begin{quote}\small\ttfamily
QUESTION:\\ \{qblock\}\\[2pt]
NOTES SO FAR:\\ \{notes\}\\[2pt]
How confident are you (0-100) that the notes are sufficient to answer correctly? Reply with ONLY a number.\\
Confidence:
\end{quote}

\textbf{END gate.}
\begin{quote}\small\ttfamily
QUESTION:\\ \{qblock\}\\[2pt]
NOTES SO FAR:\\ \{notes\}\\[2pt]
Decide whether the notes contain enough information to answer the question. ONLY when enough information is collected, return <next>end</next>. Otherwise return <next>continue</next>.\\
Decision:
\end{quote}








\section{LongBench-v2 by Domain}\label{app:domains}

\begin{table}[h]
\centering
\small
\caption{\method{} token savings on LongBench-v2 by domain at each model's best $\theta$, ordered by mean chunks per document.}
\label{tab:domains}
\begin{tabular}{@{}l r r rr@{}}
\toprule
Domain & $n$ & Mean chunks & Saving, Qwen3.5 & Saving, Kimi \\
\midrule
Long-dialogue history      & 39  & 13.9  & 31\% & 31\% \\
Single-document QA         & 175 & 19.1  & 39\% & 28\% \\
Multi-document QA          & 125 & 22.4  & 31\% & 32\% \\
Long in-context learning   & 81  & 36.3  & 50\% & 45\% \\
Long structured data       & 33  & 51.8  & 37\% & 42\% \\
Code repository            & 50  & 150.2 & 62\% & 67\% \\
\bottomrule
\end{tabular}
\end{table}

Per-domain accuracy differences between \method{} and full reading are small and inconsistent across the two models, so we do not interpret them as domain-level accuracy effects. The clearest efficiency result is on code repositories, which have by far the longest contexts at 150.2 chunks on average and yield the largest savings: 62\% on Qwen3.5 and 67\% on Kimi.

\section{Sensitivity}\label{app:grid}

\begin{table}[h]
\centering
\small
\caption{
Accuracy and mean token cost on S-NIAH, Qwen3-14B, all 250 questions, as
$\theta$ varies. Every $\epsilon\in\{0.005,0.01,0.02,0.05\}$ produces the
same results.
}
\label{tab:sniah-threshold-grid}
\setlength{\tabcolsep}{7pt}
\begin{tabular}{@{}lrrrrrrrr@{}}
\toprule
$\theta$
& 0.5 & 0.8 & 0.9 & 0.95 & 0.97 & 0.98 & 0.99 & 0.995 \\
\midrule
Acc.
& 0.764 & 0.788 & 0.856 & 0.872
& 0.900 & 0.912 & 0.956 & \textbf{0.980} \\

Tokens (K)
& 26.1 & 27.1 & 31.1 & 31.8
& 33.1 & 33.6 & 36.5 & 37.7 \\
\bottomrule
\end{tabular}
\end{table}

\textbf{Confidence scales.}
The highest-accuracy LongBench-v2 operating points, $\theta=0.995$ for Qwen3.5 and $\theta=0.92$ for Kimi, lie at similar positions in their respective full-reading confidence distributions: the 54th and 55th percentiles, respectively, despite median confidences of 0.992 and 0.873. This similarity does not yield a general percentile-based rule. On S-NIAH, the corresponding percentile maps to 1.000 for models whose confidence
saturates, while for Qwen3-32B it gives $\theta=0.942$, which stops before the evidence on 30\% of questions. Thus, normalizing $\theta$ by a model's full-reading confidence distribution does not replace the conservative shared threshold used in the main experiments.

\section{Robustness of the Shared Operating Point} \label{app:operating-point-selection}

We examine whether a single operating point can be used across model families
without per-model tuning. The analysis uses the complete S-NIAH trajectories:
250 questions for each of five models. Because all models are evaluated on the
same question IDs, we apply one deterministic split shared across models.
Questions whose integer MD5 hash is even form the development set
($n=135$), and the remaining questions form the held-out test set
($n=115$).

On the development split, we search
\[
\theta \in
\{0.50,0.60,0.70,0.80,0.90,0.92,0.94,0.95,
  0.96,0.97,0.98,0.99,0.995\}
\]
and
\[
\varepsilon\in\{0.005,0.01,0.02,0.05\},
\]
with the stability window fixed at $w=3$. For each model, let
$A^\star_{\mathrm{dev}}$ denote the highest development accuracy observed on
the grid. We then select the lowest-cost configuration satisfying
\[
A_{\mathrm{dev}}\geq A^\star_{\mathrm{dev}}-0.02,
\]
breaking ties by higher development accuracy and then by the more conservative
threshold. This produces a model-specific cost-aware operating point that we
compare against the shared fixed configuration
$(\theta,\varepsilon,w)=(0.995,0.05,3)$ on the same held-out questions.

\begin{table}[H]
\centering
\small
\caption{
Model-specific cost-aware tuning versus the shared fixed operating point on
S-NIAH. All results use the same 115-question held-out split.
The shared configuration is
$(\theta,\varepsilon,w)=(0.995,0.05,3)$.
}
\label{tab:sniah-operating-point-selection}
\setlength{\tabcolsep}{4.5pt}
\begin{tabular}{@{}lcccc@{}}
\toprule
Model
& Tuned $(\theta,\varepsilon)$
& Tuned Acc.
& Fixed Acc.
& Fixed Premature $\downarrow$ \\
\midrule
Qwen2.5-7B  & $(0.98,0.05)$  & 0.878 & 0.922 & 6.1\%  \\
Qwen3-14B   & $(0.995,0.05)$ & 0.965 & 0.965 & 2.6\%  \\
Qwen3-32B   & $(0.80,0.05)$  & 0.965 & 1.000 & 0.0\%  \\
Gemma-3-12B & $(0.995,0.05)$ & 0.983 & 0.983 & 1.7\%  \\
Gemma-3-27B & $(0.50,0.02)$  & 0.791 & 0.835 & 16.5\% \\
\midrule
\textbf{Mean}
& ---
& \textbf{0.917}
& \textbf{0.941}
& \textbf{5.4\%} \\
\bottomrule
\end{tabular}
\end{table}

The cost-aware tuned operating points vary substantially across models, from $\theta=0.50$ to $0.995$, whereas the shared conservative configuration retains equal or higher held-out accuracy on every model and yields a higher mean held-out accuracy, 0.941 versus 0.917. This supports the robustness of the fixed $\theta=0.995$ operating point across models.

\section{Additional Results}\label{app:more}

\begin{table}[H]
\centering
\small
\caption{Probe-overhead accounting on LongBench-v2, all 503 questions. Savings are relative to full reading: 382{,}971 tokens for Qwen3.5 and 340{,}761 for Kimi. ``Without probes'' is a counterfactual accounting ablation that preserves the same stopping decisions and accuracy while removing intermediate probe calls; the final probe used to produce the answer is retained.}
\label{tab:longbench-probe-overhead}
\setlength{\tabcolsep}{4pt}
\begin{tabular}{@{}llrrrrr@{}}
\toprule
Model & Configuration & Acc. & With probes & Probe cost
& Without probes & Saving w/o probes \\
\midrule
Qwen3.5 & fixed/best
& 0.541 & 190{,}966 & 24{,}079 & 166{,}887 & 56.4\% \\
Kimi & fixed
& 0.519 & 331{,}070 & 37{,}988 & 293{,}082 & 14.0\% \\
Kimi & best $\theta$
& 0.535 & 172{,}648 & 19{,}968 & 152{,}680 & 55.2\% \\
\bottomrule
\end{tabular}
\end{table}

Probe calls introduce a measurable token cost, but this overhead is already included in all main results; removing it counterfactually would increase the available savings to 56.4\% on Qwen3.5 and up to 55.2\% on Kimi.

\begin{table}[H]
\centering
\small
\caption{
Threshold sweep on RULER-HotpotQA, all 500 trajectories,
$\epsilon=0.05$, $w=3$. Pre-proxy stopping is measured on the 455 trajectories with a located nontrivial literal answer mention. Accuracy uses the official RULER substring criterion, and saving is relative to full reading.
}
\label{tab:ruler-threshold-sweep}
\setlength{\tabcolsep}{4pt}
\begin{tabular}{@{}crrrrrrrrrr@{}}
\toprule
& \multicolumn{2}{c}{Qwen3.5}
& \multicolumn{2}{c}{Kimi}
& \multicolumn{2}{c}{Qwen3-14B}
& \multicolumn{3}{c}{Premature} \\
\cmidrule(lr){2-3}
\cmidrule(lr){4-5}
\cmidrule(lr){6-7}
\cmidrule(l){8-10}
$\theta$
& Acc. & Saving
& Acc. & Saving
& Acc. & Saving
& Qwen3.5 & Kimi & Qwen3-14B \\
\midrule
0.80
& 0.660 & 36\%
& 0.730 & 32\%
& 0.532 & 53\%
& 10.3\% & 7.9\% & 20.7\% \\

0.90
& 0.682 & 31\%
& 0.742 & 26\%
& 0.556 & 48\%
& 8.4\% & 6.6\% & 18.1\% \\

0.92
& 0.686 & 30\%
& 0.750 & 24\%
& 0.556 & 48\%
& 7.7\% & 5.1\% & 17.7\% \\

0.95
& 0.698 & 26\%
& 0.752 & 19\%
& 0.562 & 46\%
& 6.2\% & 3.5\% & 16.3\% \\

0.98
& 0.708 & 21\%
& 0.754 & 11\%
& 0.580 & 40\%
& 4.0\% & 1.8\% & 12.2\% \\

0.99
& \textbf{0.714} & \textbf{19\%}
& 0.756 & 6\%
& 0.590 & 37\%
& 3.3\% & 1.3\% & 11.0\% \\

0.995
& \textbf{0.714} & 16\%
& \textbf{0.758} & \textbf{3\%}
& \textbf{0.608} & \textbf{32\%}
& \textbf{2.4\%} & \textbf{0.7\%} & \textbf{8.8\%} \\

\midrule
\rowcolor{gray!15}
full reading
& 0.722 & 0\%
& 0.760 & 0\%
& 0.646 & 0\%
& 0.0\% & 0.0\% & 0.0\% \\
\bottomrule
\end{tabular}
\end{table}

Across all three models, increasing $\theta$ reduces pre-proxy stopping and moves accuracy toward full reading at the cost of lower savings. Qwen3.5 reaches its highest accuracy at $\theta=0.99$, while Kimi and Qwen3-14B continue improving through $\theta=0.995$. The effect is especially pronounced for Qwen3-14B, where the stricter threshold reduces premature stopping from 20.7\% to 8.8\% while increasing accuracy from 0.532 to 0.608.

\begin{table}[H]
\centering
\small
\caption{
LongBench-v2 chunk-size and notes-cap ablations with Kimi K2.5 on the prespecified 80-question subset, using $\theta=0.92$, $\epsilon=0.05$, and $w=3$. Regret is measured relative to full reading under the same \(L\) or \(B\) configuration.
}
\label{tab:longbench-chunk-notes-ablation}
\setlength{\tabcolsep}{7pt}
\begin{tabular}{@{}llrrr@{}}
\toprule
Axis & Configuration & Acc. & Regret $\downarrow$ & Tokens \\
\midrule

\multirow{3}{*}{Chunk size $L$}
& 12K
& 0.425 & 3.8\% & 209{,}470 \\

& 24K (default)
& \textbf{0.538} & \textbf{1.3\%} & 168{,}160 \\

& 48K
& 0.500 & \textbf{1.3\%} & \textbf{151{,}960} \\

\midrule

\multirow{3}{*}{Notes cap $B$}
& 3K
& 0.525 & 2.5\% & \textbf{98{,}755} \\

& 6K (default)
& \textbf{0.538} & \textbf{1.3\%} & 168{,}160 \\

& 12K
& 0.475 & 2.5\% & 188{,}747 \\

\bottomrule
\end{tabular}
\end{table}

On this LongBench-v2 subset, the default $L=24$K and $B=6$K give the highest accuracy, while moving either parameter exposes different accuracy--cost trade-offs rather than improving both simultaneously.

\begin{table}[H]
\centering
\small
\caption{Stability-window ablation on LongBench-v2, all 503 questions.
Qwen3.5 uses $\theta=0.995$ and Kimi uses $\theta=0.92$;
$\varepsilon=0.05$. Saving is relative to full reading.}
\label{tab:longbench-window}
\setlength{\tabcolsep}{5pt}
\begin{tabular}{@{}crrr rrr@{}}
\toprule
& \multicolumn{3}{c}{Qwen3.5-397B-A17B}
& \multicolumn{3}{c}{Kimi K2.5} \\
\cmidrule(lr){2-4}\cmidrule(l){5-7}
$w$ & Acc. & Tokens & Saving & Acc. & Tokens & Saving \\
\midrule
2 & 0.533 & 185{,}722 & 51.5\% & 0.529 & 168{,}644 & 50.5\% \\
\textbf{3 (default)}
  & \textbf{0.541} & \textbf{190{,}966} & \textbf{50.0\%}
  & \textbf{0.535} & \textbf{172{,}648} & \textbf{49.0\%} \\
4 & 0.537 & 202{,}054 & 47.2\% & 0.533 & 180{,}912 & 46.9\% \\
5 & \textbf{0.541} & 206{,}716 & 46.0\% & 0.527 & 194{,}035 & 43.1\% \\
\midrule
\rowcolor{gray!15}
full reading & 0.535 & 382{,}971 & 0\%
             & 0.517 & 340{,}761 & 0\% \\
\bottomrule
\end{tabular}
\end{table}

Accuracy varies by at most 0.8 points across $w=2$--$5$ on either model. The default $w=3$ gives the highest accuracy on Kimi and ties for the highest on Qwen3.5, while preserving greater savings than the larger windows.

\begin{table*}[h]
\centering
\small
\caption{
Per-model verbalized-threshold ablation on S-NIAH, with 250 questions per
model. Thresholds rescale the verbalized 0--100 output to $[0,1]$.
The selected $0.995$ setting is the strictest point on the prespecified grid.
}
\label{tab:verbalized-threshold-per-model}
\setlength{\tabcolsep}{5pt}
\begin{tabular}{@{}llrrrrrrr@{}}
\toprule
Model & Metric
& $0.80$ & $0.90$ & $0.92$ & $0.95$
& $0.98$ & $0.99$ & $\mathbf{0.995}$ \\
\midrule

\multirow{2}{*}{Qwen2.5-7B}
& Acc.
& 0.568 & 0.596 & 0.596 & 0.596 & 0.700 & 0.700 & \textbf{0.700} \\
& Premature $\downarrow$
& 43.2\% & 40.4\% & 40.4\% & 40.4\% & 29.2\% & 29.2\% & \textbf{29.2\%} \\
\addlinespace

\multirow{2}{*}{Qwen3-14B}
& Acc.
& 0.916 & 0.916 & 0.916 & 0.916 & 0.916 & 0.916 & \textbf{0.916} \\
& Premature $\downarrow$
& 8.4\% & 8.4\% & 8.4\% & 8.4\% & 8.4\% & 8.4\% & \textbf{8.4\%} \\
\addlinespace

\multirow{2}{*}{Qwen3-32B}
& Acc.
& 0.588 & 0.592 & 0.592 & 0.592 & 0.592 & 0.592 & \textbf{0.592} \\
& Premature $\downarrow$
& 41.2\% & 40.8\% & 40.8\% & 40.8\% & 40.8\% & 40.8\% & \textbf{40.8\%} \\
\addlinespace

\multirow{2}{*}{Gemma-3-12B}
& Acc.
& 0.544 & 0.544 & 0.544 & 0.544 & 0.544 & 0.544 & \textbf{0.544} \\
& Premature $\downarrow$
& 45.6\% & 45.6\% & 45.6\% & 45.6\% & 45.6\% & 45.6\% & \textbf{45.6\%} \\
\addlinespace

\multirow{2}{*}{Gemma-3-27B}
& Acc.
& 0.808 & 0.816 & 0.840 & 0.840 & 0.876 & 0.876 & \textbf{0.876} \\
& Premature $\downarrow$
& 18.8\% & 18.0\% & 15.6\% & 15.6\% & 12.0\% & 12.0\% & \textbf{12.0\%} \\

\bottomrule
\end{tabular}
\end{table*}

Raising the verbalized threshold helps some models but has little or no effect
on others. We therefore use $0.995$, the strictest prespecified value, as a
single conservative threshold across models. It attains or ties the best
observed operating point for every model, yet verbalized stopping remains highly
premature on Qwen3-32B and Gemma-3-12B, showing that threshold adjustment alone
does not resolve its model dependence.

\end{document}